%% file: main.tex
\documentclass{article}
\usepackage{jfrExamplee}
\usepackage{graphicx}
\usepackage{apalike}
\usepackage{setspace}

\usepackage{amsmath} 
\usepackage{amssymb}  
\usepackage{graphicx}
\usepackage{subfig}
\usepackage{multirow}
\usepackage{amsfonts}
\usepackage{hyperref}
\usepackage{tabularx}
\usepackage{algorithmic}
\usepackage{algorithm}
\usepackage{siunitx}
\usepackage{url}
\usepackage{bm}
\usepackage{breakurl}
\usepackage{enumerate}
\usepackage[usenames,dvipsnames]{color}

\newcommand{\fig}{Fig.~}

\usepackage{verbatim}
\usepackage{tikz}
\usepackage{pgfplots}
\usepackage{pgfgantt}
\usetikzlibrary{backgrounds,arrows,automata,shapes,positioning,calc,through}
\pgfdeclarelayer{background}
\pgfdeclarelayer{foreground}
\pgfsetlayers{background,main,foreground}
\usepackage{rotating}

\newcommand{\todoi}[1]{{\color{blue}{ #1}}}

\title{\LARGE \bf Design and Deployment of an Autonomous Unmanned Ground Vehicle  for Urban Firefighting Scenarios}

\author{
Kshitij Jindal\\
Department of MAE\\
New York University\\
Brooklyn, 11201 \\
\texttt{kshitij.jindal@nyu.edu} \\
\And
Anthony Wang \\
Department of MAE\\
New York University\\
Brooklyn, 11201 \\
\texttt{aw3645@nyu.edu} \\
\And
Dinesh Thakur \\
University of Pennsylvania \\
3330 Walnut Street \\
Philadelphia, 19104 \\
\texttt{tdinesh@seas.upenn.edu} \\
\And
Alex Zhou\\
University of Pennsylvania \\
3330 Walnut Street \\
Philadelphia, 19104 \\
\texttt{alexzhou@seas.upenn.edu} \\
\And
Vojtech Spurny\\
Faculty of Electrical Engineering\\
CTU in Prague\\
Prague
166 27, Czech Republic\\
\texttt{vojtech.spurny@fel.cvut.cz}
\And
Viktor Walter\\
Faculty of Electrical Engineering\\
CTU in Prague\\
Prague
166 27, Czech Republic\\
\texttt{waltevik@fel.cvut.cz}
\And
George Broughton\\
Faculty of Electrical Engineering\\
CTU in Prague\\
Prague
166 27, Czech Republic\\
\texttt{george.broughton@fel.cvut.cz}
\And
Tomas Krajnik\\
Faculty of Electrical Engineering\\
CTU in Prague\\
Prague
166 27, Czech Republic\\
\texttt{tomas.krajnik@fel.cvut.cz}
\AND
Martin Saska\\
Faculty of Electrical Engineering\\
CTU in Prague\\
Prague
166 27, Czech Republic\\
\texttt{saska1@fel.cvut.cz}
\And
Giuseppe Loianno\\
Department of ECE and MAE\\
New York University\\
Brooklyn, 11201 \\
\texttt{loiannog@nyu.edu} \\
}

\begin{document}  
    
\maketitle

\begin{abstract}
Autonomous mobile robots have the potential to solve missions that are either too complex or dangerous to be accomplished by humans. In this paper, we address the design and autonomous deployment of a ground vehicle equipped with a robotic arm for urban firefighting scenarios.
We describe the hardware design and algorithm approaches for autonomous navigation, planning, fire source identification and abatement in unstructured urban scenarios. The approach employs on-board sensors for autonomous navigation and thermal camera information for source identification. A custom electro--mechanical pump is responsible to eject water for fire abatement. The proposed approach is validated through several experiments, where we show the ability to identify and abate a sample heat source in a building.
The whole system was developed and deployed during the Mohamed Bin Zayed International Robotics Challenge (MBZIRC) 2020, for Challenge No. 3 – Fire Fighting Inside a High-Rise Building and during the Grand Challenge where our approach scored the highest number of points among all UGV solutions and was instrumental to win the first place.  
\end{abstract}

\input{sections/introduction.tex}

\input{sections/overview.tex}

\input{sections/autonomous_navigation.tex}

\input{sections/state_estimation.tex}

\input{sections/experimental_results.tex}

\input{sections/conclusion.tex}

\bibliographystyle{apalike}
\bibliography{bibliography}

\end{document}

%% file: sections/introduction.tex
\section{Introduction}\label{sec:Introduction}

\begin{figure}[!b]
    \vspace{-10pt}
    \centering
    \includegraphics[width=0.5\linewidth]{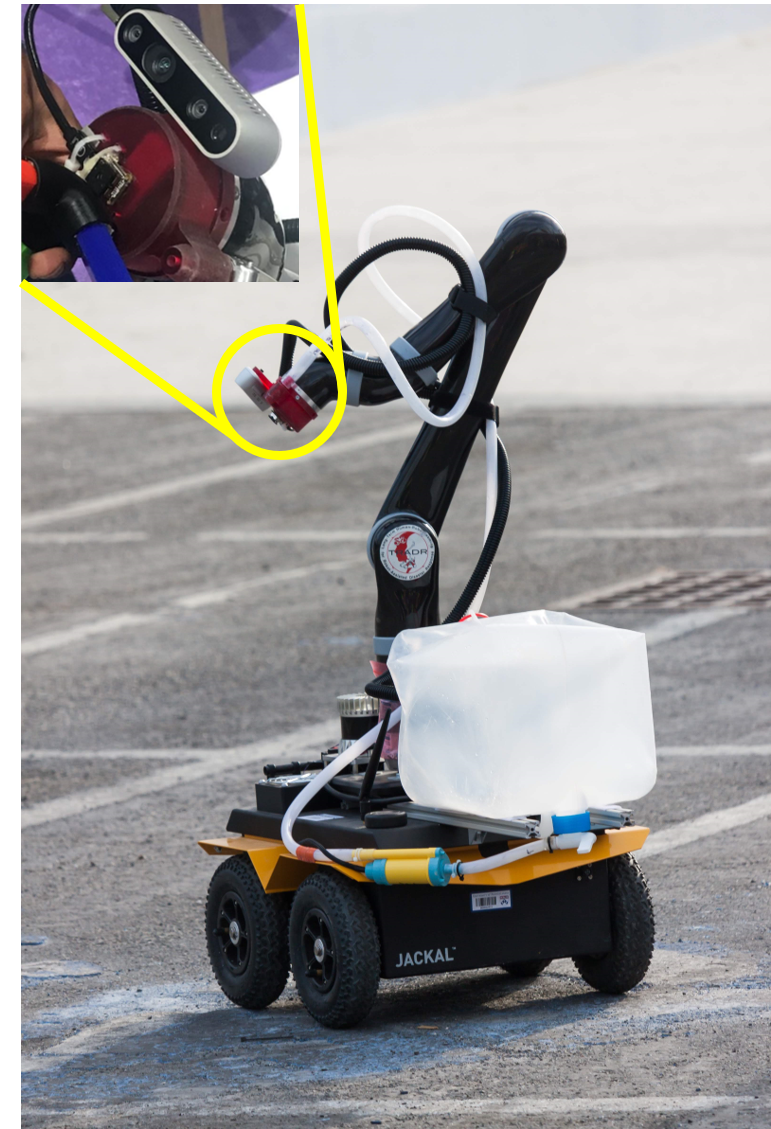}
    \caption{Proposed UGV system and manipulator for firefighting, deployed during the MBZIRC competition.}
    \label{fig:armCandid}
    \vspace{-20pt}
\end{figure}

Arm manipulators have proved to assist humans and substantially speed up tasks ranging from automotive \cite{8384639,8274801,8383972,9087500} to pharmaceutical \cite{8007498,8686979,8399114}, while increasing human safety. Unmanned Ground Vehicles (UGVs) have been employed to autonomously solve complex autonomous missions in dark or poorly illuminated confined and cluttered environments~\cite{LAMP,1242260}. A good overview and summary about the different approaches over the last 10 years is provided in~\cite{doi:10.1002/rob.21871}. The use of a mobile base coupled with a robot arm can provide the flexibility to the arm system to solve more complex tasks and to be conveniently deployed in multiple locations increasing the arm workspace.
The complex configurations, achieved by the robotic arm, aid the mobile base immensely. For instance, by attaching a camera at the end--effector of the arm, and moving it along its base joint, a 360$^{\circ}$ vision is obtained from the camera, without affecting the UGV motion. The robotic arm holds the advantage of achieving complex configurations, which aids the mobile robot base to keep its movement practical and simple. A mobile robotic arm presents the mechanical redundancy easing the detection of objects in all the robot's directions.

In 2018, only in the US, there were an estimated $379,600$ fires, which resulted in an estimated $\$8.1$ billion loss in residential buildings~\cite{residentialBuldingStats}, and $103,600$ fires, which resulted in a $\$2.6$ billion loss in non--residential buildings~\cite{nonResidentialBuldingStats}.
The reports~\cite{residentialBuldingStats,nonResidentialBuldingStats,sprinklers} indicate that systems capable of early fire suppression can reduce fire damage by more than $55$\% and life loss by $97$\%.
In particular, automatic sprinkler systems can extinguish local fires by activating only a few nozzles, limiting the fire spread as well as the damage caused by water.
However, their installation is expensive and they require maintenance and regular checks.
Moreover, traditional sprinkler heads act upon a larger area instead of delivering the water in a targeted way, causing unnecessary damage, waste of water, and sometimes failing to reach the fire~\cite{sprinklers}. 
These disadvantages can be mitigated by the use of small mobile robots, capable of early, autonomous and targeted fire suppression.

This motivates the need to create autonomous robots solutions capable of early intervention and human assistance in firefighting tasks, which are typical in urban scenarios.
This is as well the main task to solve in the Challenge 3 and in the Grand Challenge of the Mohammed Bin Zayed International Robotics Challenge\footnote{\url{http://mbzirc.com}} (MBZIRC) 2020. The robot needs to detect and extinguish a fire in an indoor environment represented by a heat source. To address this problem, we designed an autonomous mobile platform, as shown in Fig.~\ref{fig:armCandid}. The proposed system is incorporated with multiple sensors and actuators to concurrently assist the robot in navigation, heat source detection, and fire abating.



The use of robots for firefighting operations has been much discussed in the recent years. Most current firefighting robots that are deployed are teloperated. The authors in~\cite{5353970} introduce a lightweight portable robot which gathers information of the affected building and transmits it to a human controller. This method requires a human observer to control and navigate it in the environment. Furthermore, it lacks the capability to extinguish a fire. In~\cite{KHOON20121145}, the authors incorporate a flame sensor in its design. The major drawback in their method is that the mobile robot has to rotate 360$^{\circ}$ in order to scan the entire area. Moreover, the navigation system is based on a line follower method, which does not scale well to real world scenarios.
In~\cite{Zaman}, a small and cost-effective firefighting solution based on Arduino Nano is presented. The approach does not address the autonomy challenges and solutions to operate in urban firefighting scenarios. The authors focus mainly on video--streaming capabilities and the robot design is too small and with very limited computational resources to operate in urban settings.
The approach proposed in~\cite{7251507} greatly summarizes the use of a fan as an extinguishing tool in many approaches. While this strategy may work well in small scale environments to abate candles, it does not successfully scale in our envisioned scenario. Finally multiple works~\cite{Raj,8996761,4381341,7977097} present some preliminary robot designs for firefighting scenarios. However, these solutions do not show any experimental validation or deployment in real--world settings to corroborate the effectiveness of their proposed solution. 

\begin{figure}[!h]
\vspace{-10pt}
    \centering
    \includegraphics[width=0.54\linewidth]{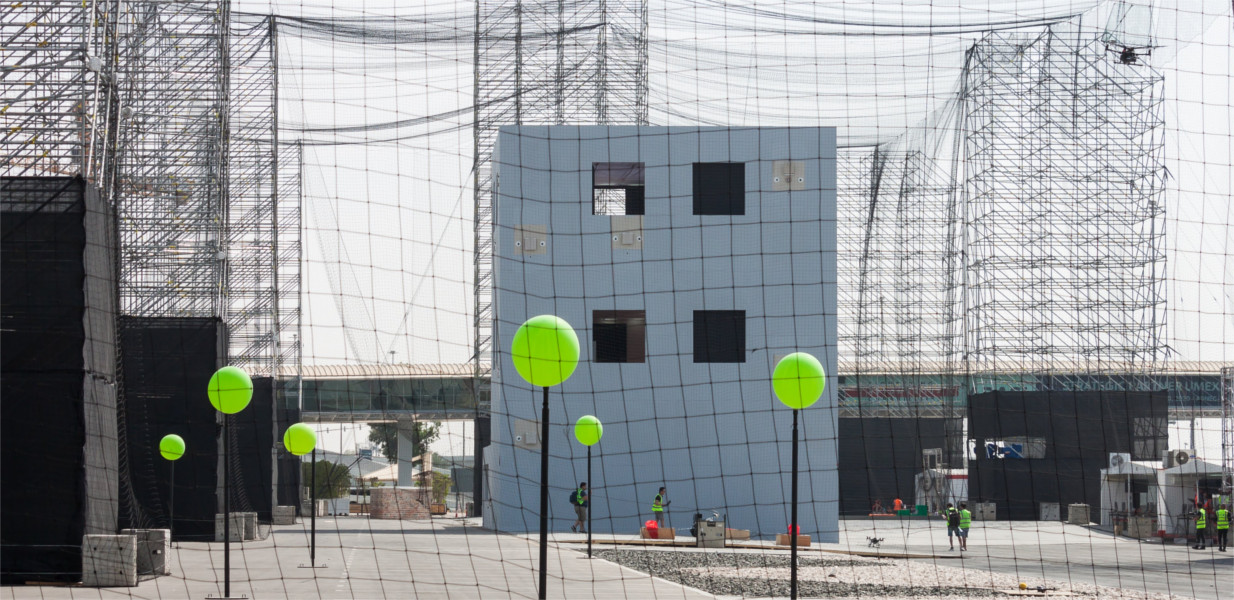}
    \includegraphics[width=0.38\linewidth]{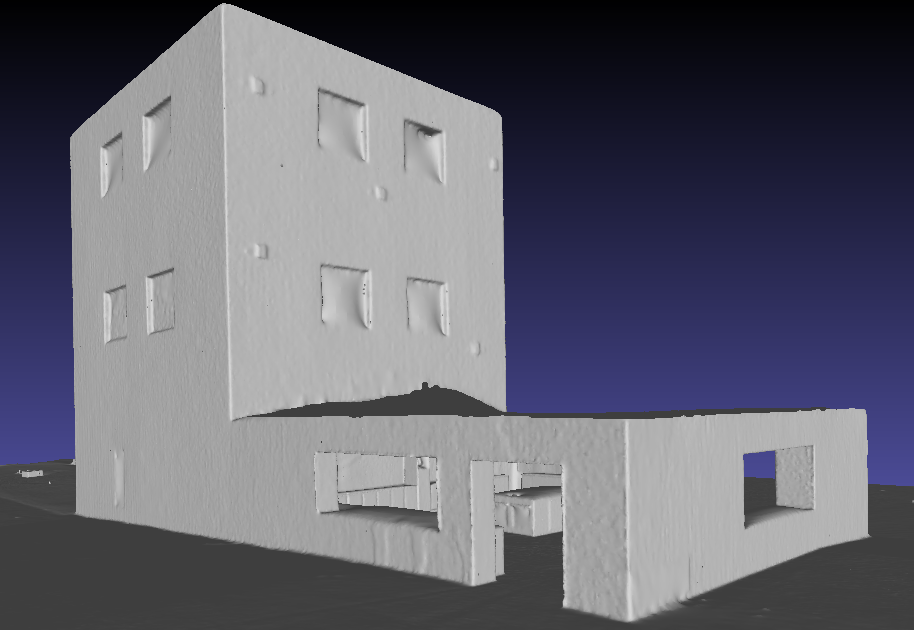}
    \caption{The tall structure simulating a high-rise building (left). A CAD of the building obtained by photogrammetry during rehearsals (right).}
    \label{fig:building_cad}
\end{figure}

In this paper, inspired by the MBZIRC 2020 Challenge No. 3 – Fire Fighting Inside a High-Rise Building as shown in Fig.~\ref{fig:building_cad} and by the competition Grand Challenge, we address the design of an autonomous ground vehicle equipped with a robotic arm, which is able to detect heat sources and abate the corresponding fire employing a water-pump system. The task outline suggested that one liter of water successfully aimed at the heat source as the goal to extinguish the synthetic fire. Our solution scored the highest number of points among all UGV firefighting solutions proposed by the 17 teams admitted to the competition during the Grand  Challenge\footnote{\url{https://www.mbzirc.com/winning-teams/2020/challenge4\#viewwinner}}, while concurrently being the most compact one. It was instrumental to win the Grand Challenge scenario.
\\

This work presents multiple contributions. 
First, we present a novel mobile platform composed by a mobile base and a robot arm. The system is equipped with a water tank used to extinguish the detected fire.
Second, we present how to obtain a reliable state estimation, planning, and control for our mobile robot solution to autonomously navigate, locate, and abate the fire sources.
Third, this is the first time that
design, estimation, planning, and control problems are addressed simultaneously
in a such challenging scenario. We test our methodologies 
in a challenging environment, where the robot needs to enter a building and solve a fire--extinguishing task. Finally, we plan to release our current
framework upon acceptance as an open source package to
the community ~\cite{github}. This work represents a first step toward robots' deployment in real world firefighting urban scenarios. The ability to test our methodologies in the MBZIRC test--bed allows us to evaluate the ability to transition our approaches to more complex settings. To contribute and boost this transition process as well as to facilitate new teams becoming part of the MBZIRC in the upcoming editions, we open--source our designs and algorithms. 


The paper is organized as follows.
Section~\ref{sec:overview} introduces the hardware design and software overview of the system.
Section~\ref{sec:autonomous_navigation} presents our methodology to control the UGV and the incorporated robotic arm.
Section~\ref{sec:aim} explains our strategy to detect the heat source and then to align the UGV as closely to the identified target.
Section~\ref{sec:experimental_results} presents our experimental results, and Section~\ref{sec:conclusion} concludes the work and presents future opportunities.
\begin{figure}[!t]
    \centering
    \includegraphics[width=0.7\linewidth]{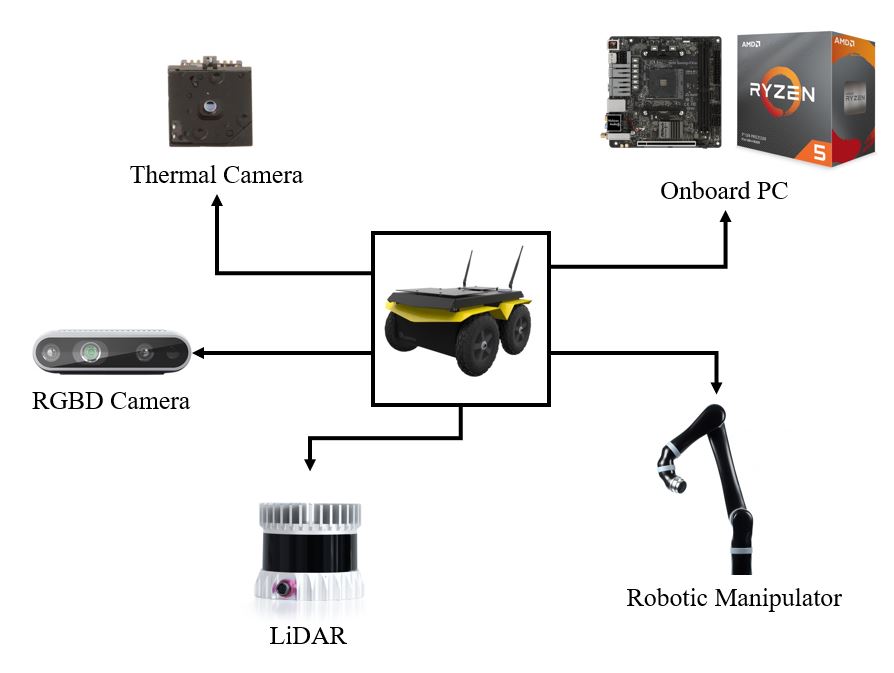}
    \caption{The proposed system architecture.}
    \label{fig:hardware}
     \vspace{-30pt}
\end{figure}

%% file: sections/overview.tex

\section{System Overview}\label{sec:overview}
\subsection{Hardware Overview}
In order to fulfill the task of extinguishing fire in an indoor environment, a fire-fighting ground robot has been designed. Figure \ref{fig:hardware} visualizes all the sensors integrated in the system independently.

The system is made of an UGV Clearpath Jackal, a 32-beam Ouster LiDAR, a 6-DoF Kinova Gen2 robotic lightweight arm mounted on the top of the ground vehicle and equipped with an RGB camera, and a GSI outdoors $15$ litre water container.
The integrated robot is shown in Fig.~\ref{fig:armCandid}. 
The Ouster LiDAR is used to generate a point cloud of the environment for localization and mapping, along with the IMU integrated with Jackal UGV.
A thermal camera is used for detecting high temperature hot--spots in the environment, and an RGB-D camera is utilized to measure its distance and position.
A water-pump system is designed and manufactured to abate the detected fire.

The maximum speed of the UGV is $2.0$ m/s, with a maximum payload of $20$ Kg. The UGV is also equipped with a 3DM-GX3-25\footnote{\url{https://www.microstrain.com/inertial/3dm-gx3-25}} IMU, that has a data output rate of up to 1000 Hz. The Kinova Gen2 Ultra lightweight robotic arm is a 6 DoF curved wrist arm with a reach of $90~\si{cm}$. It has a mid-range continuous maximum payload of $2.6$ Kg, with $2.2$ Kg full-reach peak/temporary maximum payload. It has a maximum linear arm speed of $20$ cm/s. We designed a custom end-effector, which is equipped with an Intel RealSense D435\footnote{\url{https://www.intelrealsense.com/depth-camera-d435}} depth camera and a FLIR Lepton 3.5\footnote{\url{https://lepton.flir.com/news-and-updates/new-lepton-3-5-with-radiometry}} LWIR (Long Wave Infrared) thermal camera.
The Intel RealSense D435 depth camera is utilized on the system to estimate the depth of a heat source, if present. It has a maximum range of $10\si{.m}$, with a Depth Field Of View (FOV) of $87^{\circ}\pm$3$^{\circ}$ $\times$ $58^{\circ}\pm 1^{\circ}$ $\times$ $95^{\circ}\pm 3 ^{\circ}$.
The RGB sensor FOV (H $\times$ V $\times$ D) is $69.4^{\circ}$ $\times$ $42.5^{\circ}$ $\times$ $77^{\circ}$ ($\pm$3$^{\circ}$). The RGB frame rate is $30$ fps with the depth frame rate to be up to $90$ Hz. The FLIR Lepton $3.5$ LWIR thermal, attached to the designed end-effector of the arm, has a horizontal FOV of $57^{\circ}$ and a diagonal FOV of $71^{\circ}$. The effective frame rate of the camera is $8.7$ Hz with a thermal sensitivity of 0.05$^{\circ}$C. The Ouster OS-1 $32$ channel LiDAR is used in the system for autonomous navigation. It has a maximum range of $120\si{.m}$ with a precision of $\pm(1.5-5)$ cm, a vertical FOV of $45^{\circ}$ with a vertical angular precision varying from $0.35^{\circ}$ to $2.8^{\circ}$, and a horizontal FOV of $360^{\circ}$. An AMD Ryzen 5 3600 6-Core, 12-Thread\footnote{\url{https://www.amd.com/en/products/cpu/amd-ryzen-5-3600}} with 16 GB memory is used as onboard computer to run our algorithms.

\begin{figure}[!t]
    \centering
    \includegraphics[width=1.0\linewidth]{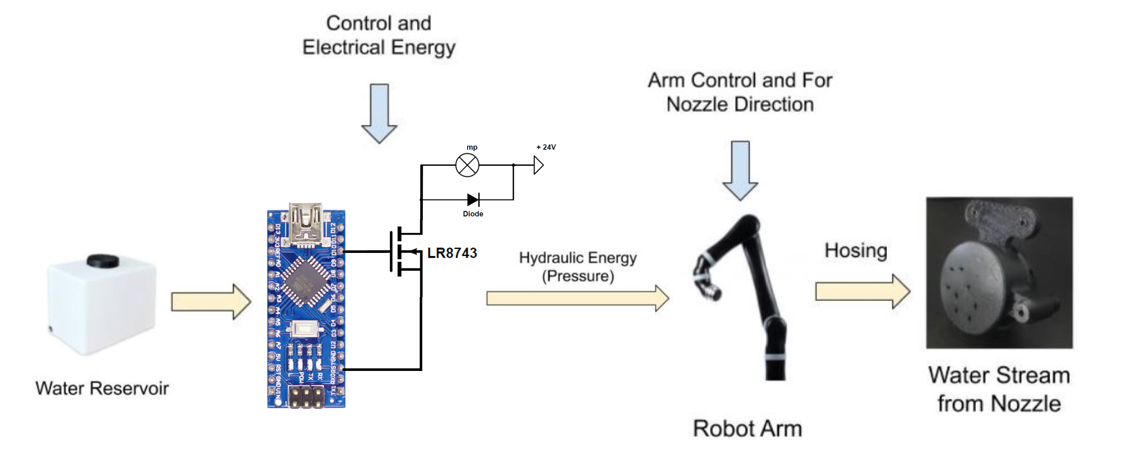}
    \caption{Flowchart of the water-pump system.}
    \label{fig:waterPumpHardware}
    \vspace{-10pt}
\end{figure}

A 15 liter water cube\footnote{\url{https://gsioutdoors.com/15-l-water-cube.html}} is located at the rear of the UGV to supply the water pump. It is controlled by our system to pressurize the water through the piping system and exits from the nozzle. The direction of the nozzle can be altered by the movement of the robot arm and the UGV platform to guide the water stream to the desired location. The water is ejected from the nozzle, housed inside the end-effector supplied from a water pump. An overview of the water projecting mechanism is demonstrated in \fig\ref{fig:waterPumpHardware}.

The constraints and performance requirements that our fire extinguishing system needs to achieve are the water stream reach and flow rate. To prevent the water from reflecting back on to the robot once hitting a hard surface, the water should reach at least $1.5$ meters. Therefore, after sizing the appropriate motor, piping system, and nozzle design, a $24$ V DC inline pump that is capable of delivering $1$ bar of pressure has been employed. The piping of the system routes the pressurized water stream from the base of the robot arm to the tip of the end-effector. The design of the end-effector is shown in the top-left corner of \fig\ref{fig:armCandid}.

\begin{figure}[!t]
    \centering
    \includegraphics[width=0.45\linewidth]{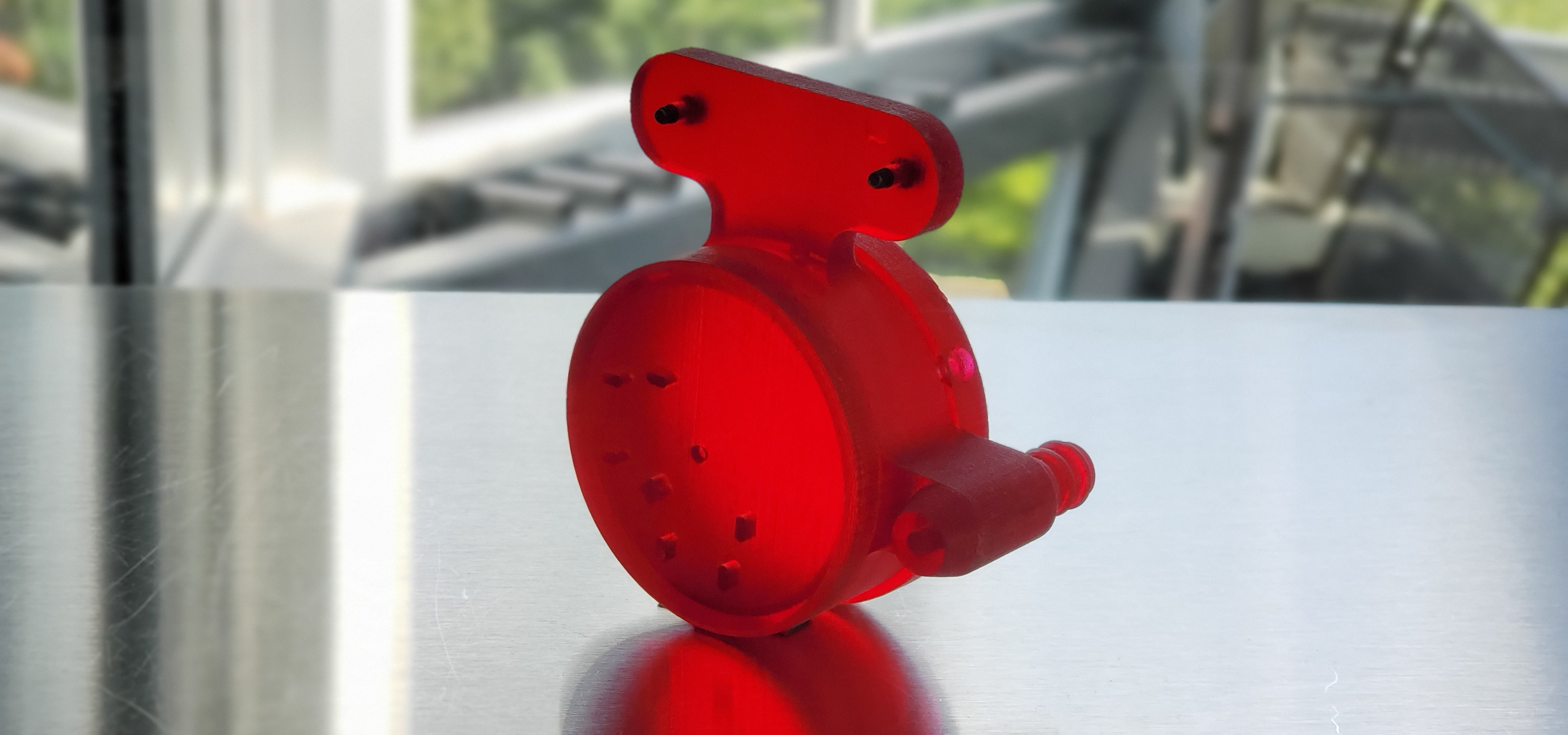}
        \includegraphics[width=0.46\linewidth]{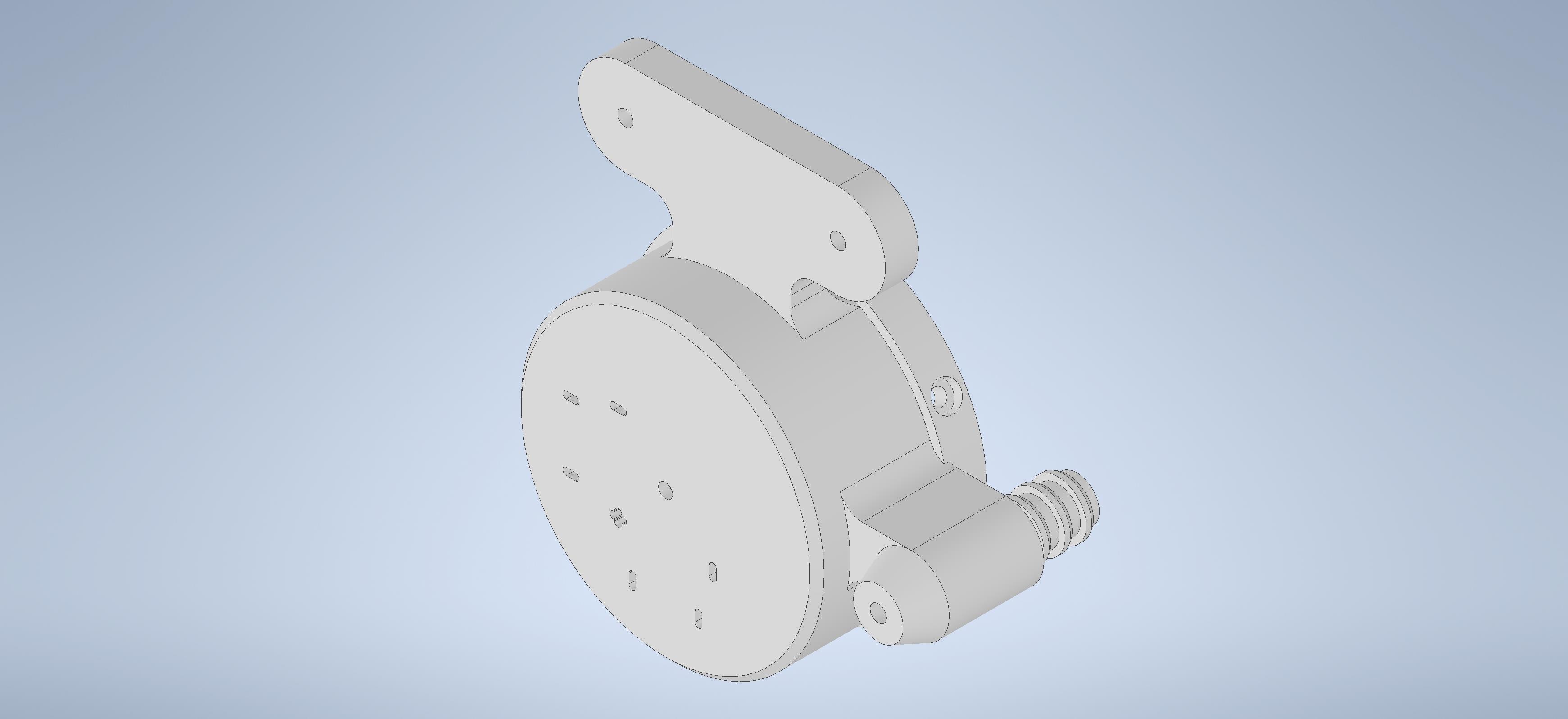}
    \caption{Designed CAD end-effector (left) and corresponding 3D printed prototype (right).}
    \label{fig:endEffector2}
    \vspace{-10pt}
\end{figure}

The end-effector is designed to house three components, a RGB-D camera, a thermal camera, and a nozzle. 
In \fig\ref{fig:endEffector2}, the flange extended on the top is used for housing the RGB-D camera, specifically designed for the Intel RealSense D435. The thermal camera is designed to mount right on the center of the circular surface and aligning with the axis of rotation of joint $6$. Lastly, the nozzle is extended from the side of the end-effector and has a male receptor for connecting the piping and an opening on the other end for the jet stream to exit from. The nozzle is $0.15$ inches in diameter which yields an exit velocity that meets our requirement.
The water pump is controlled using an Arduino Nano\footnote{\url{https://store.arduino.cc/usa/arduino-nano}}. The output of the Arduino is connected to a transistor which acts as an electronically activated switch to supply $24$ V DC to the water pump. A diode is also placed into circuit to prevent back EMF from damaging the controller. Finally, the actuation of the water pump is controlled using PWM signal. The Arduino Nano uses ATmega328 USB to TTL serial chip to communicate with the on--board PC. Figure~\ref{fig:waterPumpHardware} presents the circuit diagram to control our water pump system.

\subsection{Software Overview}
To achieve the autonomous fire detection task, the proposed system needs to accomplish different intermediate tasks such as autonomous navigation through the environment, the robot arm control, the detection of the heat sources, and the deployment of the water pump system. Using the hardware design described in the previous section, we provide an overview of the software system architecture. The implementation of all the functions relies on Robotic Operating System (ROS)~\cite{ros}, which is an open-source development tool widely used nowadays in the robotics community. Since this is a mobile system, a navigation system is needed to drive the UGV from initial state to target locations. The combination of LiDAR along with the default IMU of the jackal UGV are the incorporated sensors which help the system in achieving this goal. Further, a heat detection algorithm is required. The thermal camera is the key sensor which is used to detect the heat source. Based on the heat source information, the RGB-D camera is used to locate it with respect to the robot arm. The water pump is then activated to achieve the final goal of extinguishing the fire at identified target locations. A control and planning policy is also needed for the robotic manipulator. Finally, we do require a state--machine that is able to activate and coordinate all the different sub--tasks in a sequential or concurrent manner.

%% file: sections/autonomous_navigation.tex
\section{Autonomous Navigation and Control}\label{sec:autonomous_navigation}

The ROS navigation stack move\textunderscore base\footnote{\url{http://wiki.ros.org/move_base}} is employed for navigating the Jackal UGV.
The {\it move\textunderscore base} package utilizes costmap\textunderscore 2d\footnote{\url{http://wiki.ros.org/costmap_2d}} approach to accomplish navigation tasks.
Cost map, which aids in determining the areas in the map where the robot's motion is permitted, is calculated from an occupancy grid built from the sensory data.
The generated cost map allows the UGV to plan its own path based on the goal location, obstacle configuration and actual robot position.

For estimating the UGV location, the system must have an established localization algorithm.
For this purpose, we use Lightweight Ground-Optimized LiDAR Odometry and Mapping (LeGO-LOAM)~\cite{8594299} on the point cloud generated by the LiDAR.
LeGO-LOAM is an extended version of LiDAR Odometry and Mapping (LOAM)~\cite{Zhang-2014-7903}, which extracts edge and surface point features in a LiDAR point cloud based on the roughness of the local surface.
Then the features are re-projected on the next scan based on a motion model.
Finally, the 3D motion is estimated recursively by minimizing the sum of overall distances between point correspondences.
LeGO-LOAM introduces a procedure to determine the ground plane and distinguishes between features detected on ground plane and on obstacles.
The algorithm presents better accuracy when the Jackal IMU is used since it exploits the sensor fusion among the two sensors.
Combining the localization algorithm with the move\textunderscore base navigation stack, resulted in autonomous navigation of the robot in the tested environment.

The waypoint generation by the move\textunderscore base is achieved by specifying the goal locations from the current location.
For motion planning, the stack maintains two planners: a global planner that generates a plan for the robot to navigate avoiding obstacles and a local planner that generates velocity commands to track the global plan.
The global planner goal location can either be obtained from the user or from the planning system itself.
Once a goal location is specified, a cost map is generated such to guide the UGV from current location to the goal location.
Cost map generation is dynamic, which makes it possible for the planner to update its path based on real-time occupancy.


To control the robot arm using ROS, we need to build and setup four different modules~\cite{mobile_MoveIt}.
First ROS needs to understand the modeling properties of our robot arm as well as the kinematic relationships between joints.
This is conventionally defined in robot\textunderscore description on the parameter server.
Next, we need to utilize a motion planning algorithm to compute the optimal path for the robot to reach given goals.
We determined that RRT-Connect~\cite{844730} in the Open Motion Planning Library (OMPL) is a suitable planner for our application as it presents a simple and efficient randomized algorithm for generating paths in high-dimensional configuration spaces.
Third, we need to construct an environment model, either through pre-defined user settings or interpreted by the sensors in real time.
This helps the robot arm to avoid collisions and understand where to place the end-effector to complete the mission.
Lastly, our computer needs a method to communicate with the robot arm to convert all the generated trajectories into physical motions, both in simulation and in real world.

The four modules are conveniently taken care of by Moveit!~\cite{6174325}, a motion planning library which integrates the different aspects of robot control and ensures the information consistency between each module.
By configuring it using the setup assistant GUI, we are able to load the robot properties in the form of Universal Robot Description Files (URDF).
We also configure RRT-Connect planner to be compatible with our Kinova arm so the user can provide a goal in either Cartesian or joint space, and the planner generates a trajectory for the robot arm to reach the desired location.
Additionally, communication to the robot controllers is setup to execute the planned trajectories.
The Kinova low--level driver publishes joint state information and provides a routine to execute trajectory tracking.
We can achieve the following objectives
\begin{itemize}
    \item Integrate the kinematics information of our robot arm to the motion planning library
    \item Collision checking of robot arm with the UGV, sensors and peripherals mounted on the UGV, and the environment
    \item Take into account the physical hardware limits of the specific robot arm we are using while planning and executing trajectories for task completion
    \item Dynamically update the location of the robot arm and environment
    \item Notify user when a specific motion execution is completed
\end{itemize}
All of this is done by one script file, which is launched when starting up the system.
The library contains many useful API’s we can directly or indirectly utilize to achieve the firefighting tasks such as setting goals and executing trajectories.
However, considering the re-usability of code and convenience of operation, it is desired to create a task level planner that generalizes and wraps all the useful API's into one single function.
Therefore, we created an interface that aggregates all the necessary API’s provided by Moveit! into a single action server and standardized the action goal.
Rather than writing multiple routines for path planning, users can now simply call the task-level planner using a unified message to activate the path planning pipeline.
Therefore, when controlling the robot arm in state machines, or stand-alone calls, the user can easily interface with the robot arm without having to write a separate node.







%% file: sections/state_estimation.tex
\section{Heat Source Detection and Localization}\label{sec:aim}

To extinguish the fire, the water stream has to pass through a relatively narrow aperture.
While the localization drift of the used navigation system was low, the overall positional error after travelling several meters would prevent the water stream to reach the heat source even if the heat source position would be known a--priori.
Therefore, we implemented a module, which would locate the heat source and aim the end effector precisely enough to reach the heat source.

\begin{figure}[!b]
    \centering
    \includegraphics[height=0.3\linewidth]{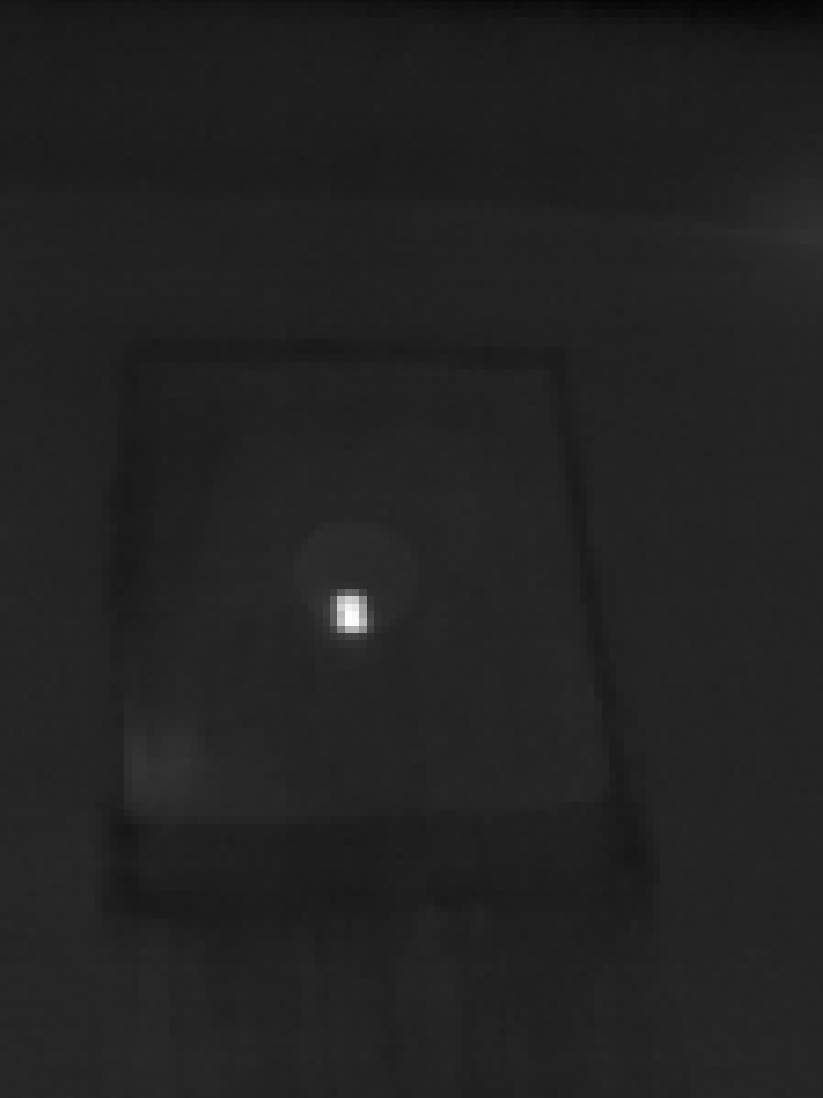}
    \includegraphics[trim=50 0 50 20, clip, height=0.3\linewidth]{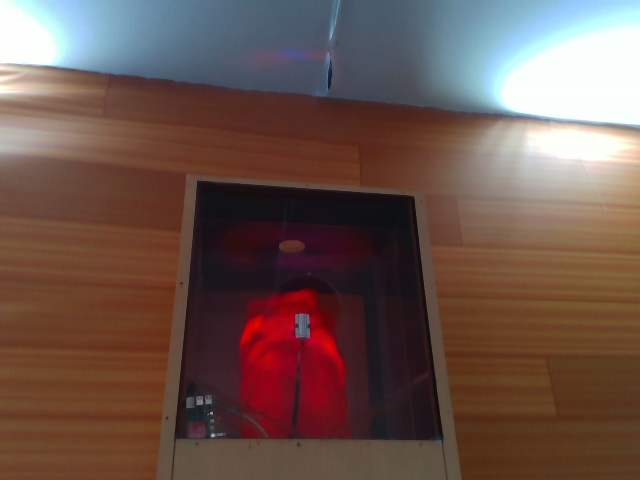}
    \includegraphics[trim=50 0 50 20, clip, height=0.3\linewidth]{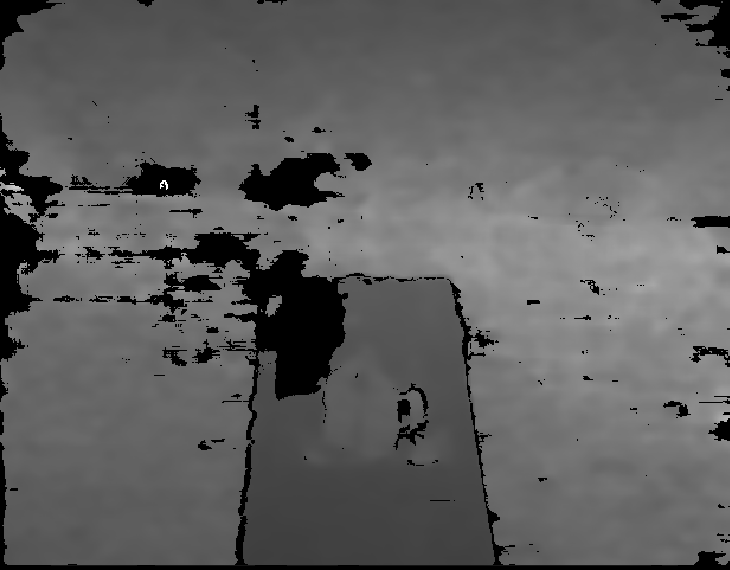}
    \caption{The white region represents the heat source present in thermal camera feed. (left). Corresponding RGB and depth images of the heat target (middle, right)}
    \label{fig:thermal2}
\end{figure}

\begin{figure}[!t]
    \centering
    \includegraphics[width=0.4\linewidth]{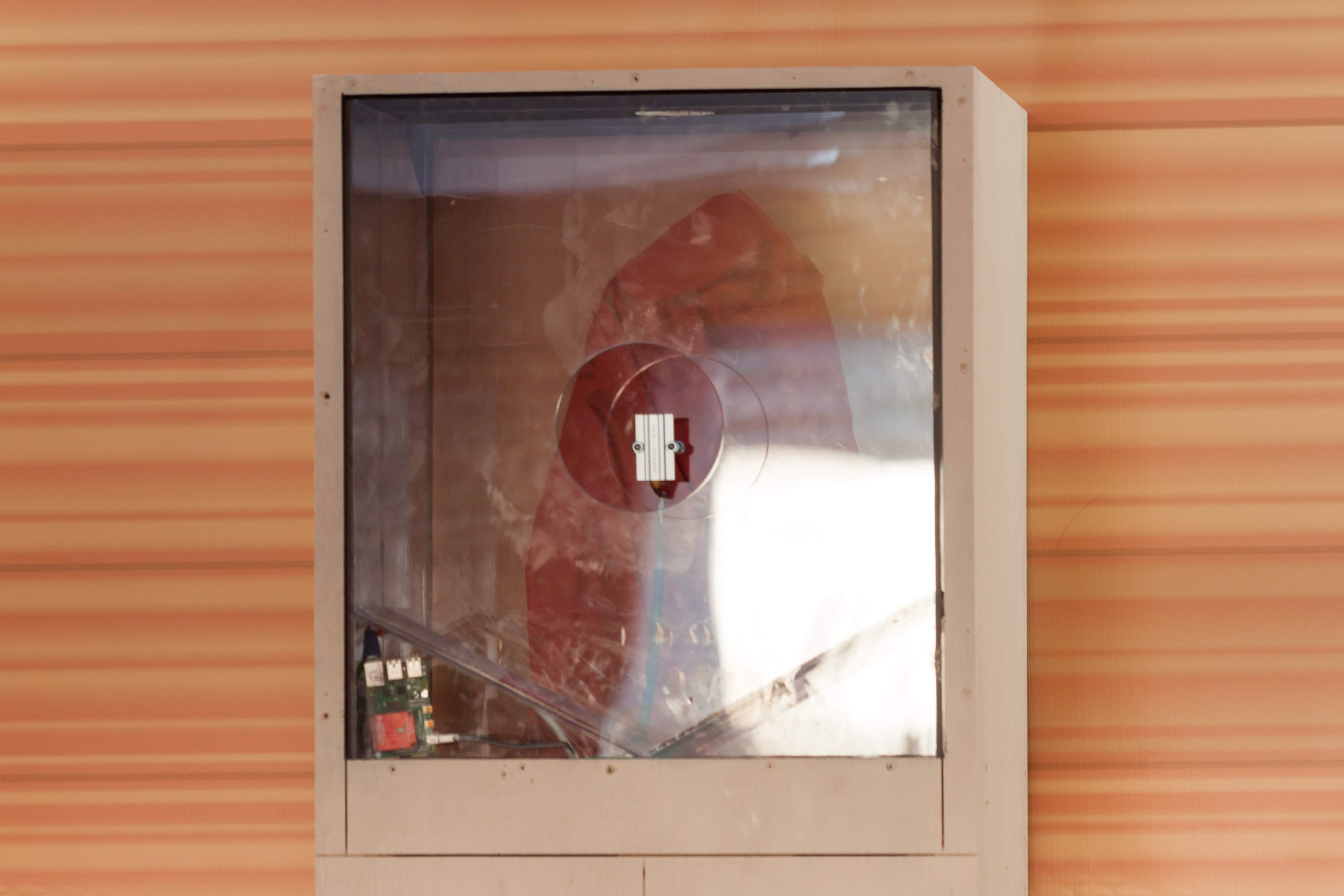}
        \includegraphics[width=0.55\linewidth]{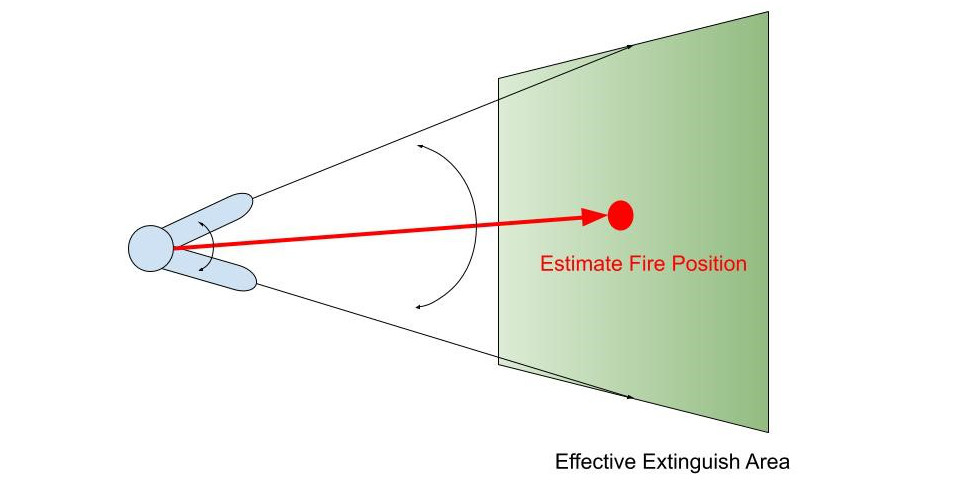}
    \caption{Image of the target utilised in the MBZIRC 2020 challenge (left) and image visualizing area covered by the water spray with the added movement (right).}
    \label{fig:target}
\end{figure}

The artificial fire sources comprised anodized aluminium heating elements of dimensions \SI{3.5}{\centi\meter} $\times$ \SI{6.0}{\centi\meter} mounted on plexiglass backplates, and were covered \SI{15}{\centi\meter} in the front by another plexiglass plate with a circular hole \SI{15}{\centi\meter} in diameter, see Fig. \ref{fig:target}.
We detected these heating elements using FLIR Lepton $3.5$ LWIR thermal camera that outputs an image where the values of pixels correspond to the estimated temperature in the given region.
The heating elements were set to maintain a temperature of \SI{120}{\celsius}, but due to the lower emissivity of anodized aluminium of $0.55$ \cite{minkina2009infrared} compared to the reference value of 1 appear to be of lower temperature, approximately \SI{70}{\celsius}.
See Fig. \ref{fig:thermal2} for an example of the view of the heat source from the thermal camera during the competition.

We assumed that the detection in this sub-task was only performed indoors where the effects of direct sunlight can be neglected and that no other objects of such temperatures will be present.
We could therefore detect the targets by simple value threshold.
Their image centers were transformed to direction vectors based on the presumed thermal camera matrix, derived from the FOV and pixel resolution of the IR camera, assuming its pinhole model.
This gave us the information on the direction of the target, which would be sufficient if the heating element was close enough to consider the water stream straight.

To reach far targets, we acquire the information on their distance by an Intel RealSense D435 depth camera attached close to the thermal camera itself with known relative transformation between the two ($5$ cm offset). Assuming pinhole camera model for the thermal camera the relative distance between the robot and the target is estimated using this ray and the depth obtained from the depth camera by re-projecting the the direction vectors of the IR camera on the depth camera. 

The precision of the distance estimate varied based on the incidence angle with respect to the front plexiglass of the target.
Since the plexiglass was transparent, sometimes the retrieved distance corresponded to the distance to the back plate instead of the front plate, or due to averaging over larger area, the distance estimate sometimes came out in between the two plates.
However, this difference was not significant for this application, since the distance of the back plate to the front one was much smaller compared to the shooting distance.
Additionally, having a line of sight to the heating element meant that we were also aiming inside of the front opening, since the plexiglass is opaque for the IR radiation band of our thermal camera.

The final output of the heating element localization module is a 3D coordinate of the heating element in the coordinate frame of the camera mounted on the end effector along with the nozzle.
This position was recalculated at the IR camera FPS, i.e. at $8$~Hz, which allowed to perform the final nozzle alignment in a closed-loop manner.
To perform both horizontal and vertical alignment we implemented two separate feedback loops controlling concurrently the robot and arm motion.

The movement of the mobile base was used to align the robot in horizontal plane. 
We used the laser measurements to align the robot with the furniture or wall with the artificial fire source and keep a distance of $1.5$~m. 
Once aligned to the wall, the robot can compensate the horizontal displacement of the fire source simply by moving forwards and backwards.
In this way, the water spray faces the artificial fire source almost perpendicularly, which maximises the cross section of the aperture it has to pass through in order to extinguish the fire.
The vertical alignment of the nozzle is performed simply by moving the third joint of the robot manipulator, which changes the nozzle pitch.
Once the vertical and horizontal alignment is finished, the water pump is turned on.

In Fig.~\ref{fig:target} (left), it can be observed that the open area of the artificial fire is rather small.
Thus, even a small misalignment or small pressure drop can cause the water stream to miss the target.
Such a situation is difficult to detect, because the water stream can obstruct the view of the artificial fire.
To avoid this, we increased the coverage of the water spray by introducing small, cyclic movements of two arm joints around the aligned position.
This results in the final spray area to form a squared section which overlaps the fire aperture.
This squared area is effectively the tolerance zone of our water spray aiming. 
The covered area obtained by adding this movement is visualized in Fig.~\ref{fig:target} (right).
In a real system, it is worth to point out that these movements would correspond to a situation, where the water is spread not only on the fire, but also around it to prevent the fire spreading. 

The different steps of the heat source detection and localization process are handled by the system state machine described in Section~\ref{sec:state_machine} and depicted in Fig.~\ref{fig:highLevelSM}.





%% file: sections/experimental_results.tex

\section{Experimental Results}\label{sec:experimental_results}
\begin{figure*}[!b]
\vspace{-10pt}
    \centering
    \includegraphics[width=0.32\textwidth]{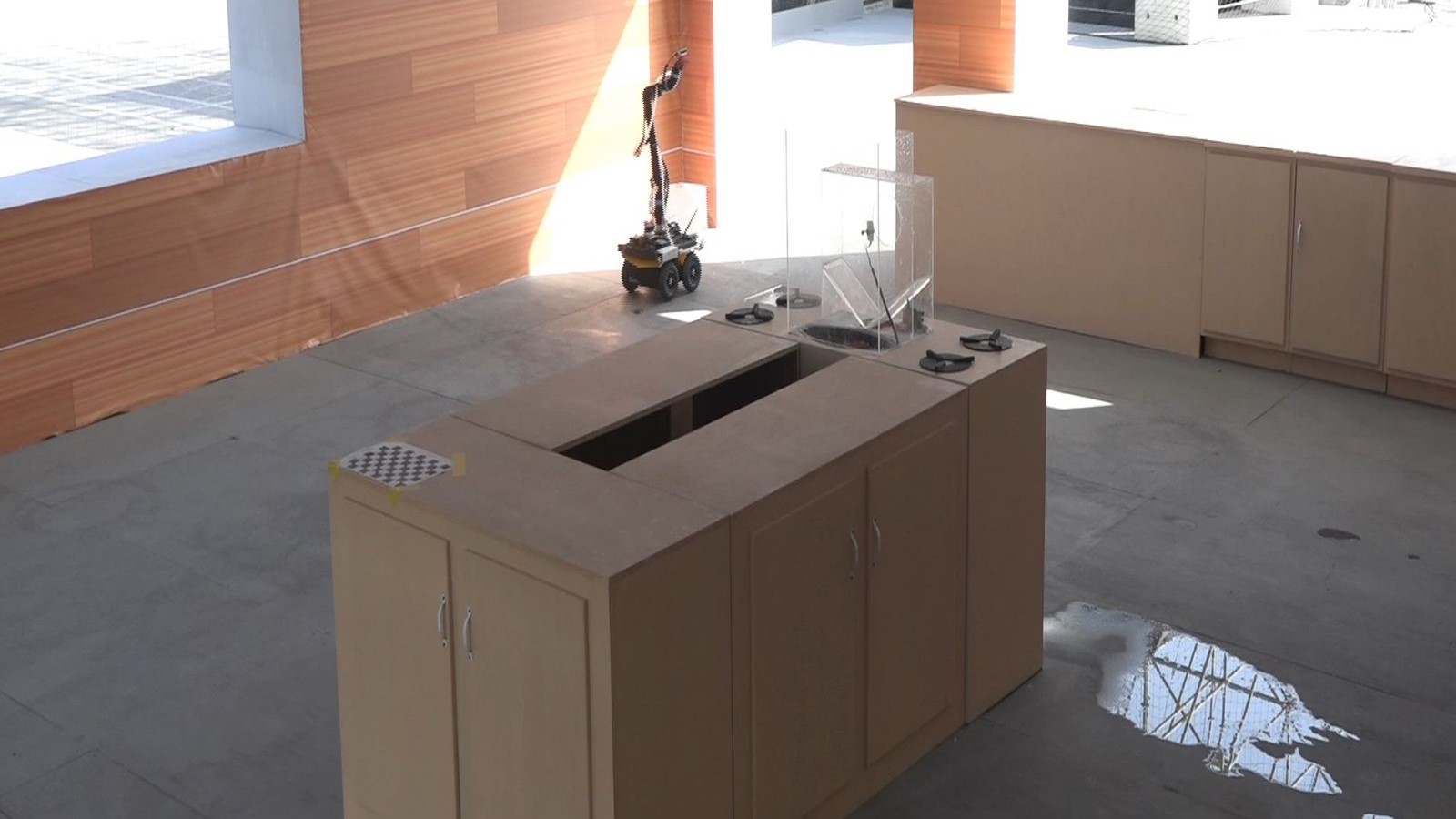}
    \includegraphics[width=0.32\textwidth]{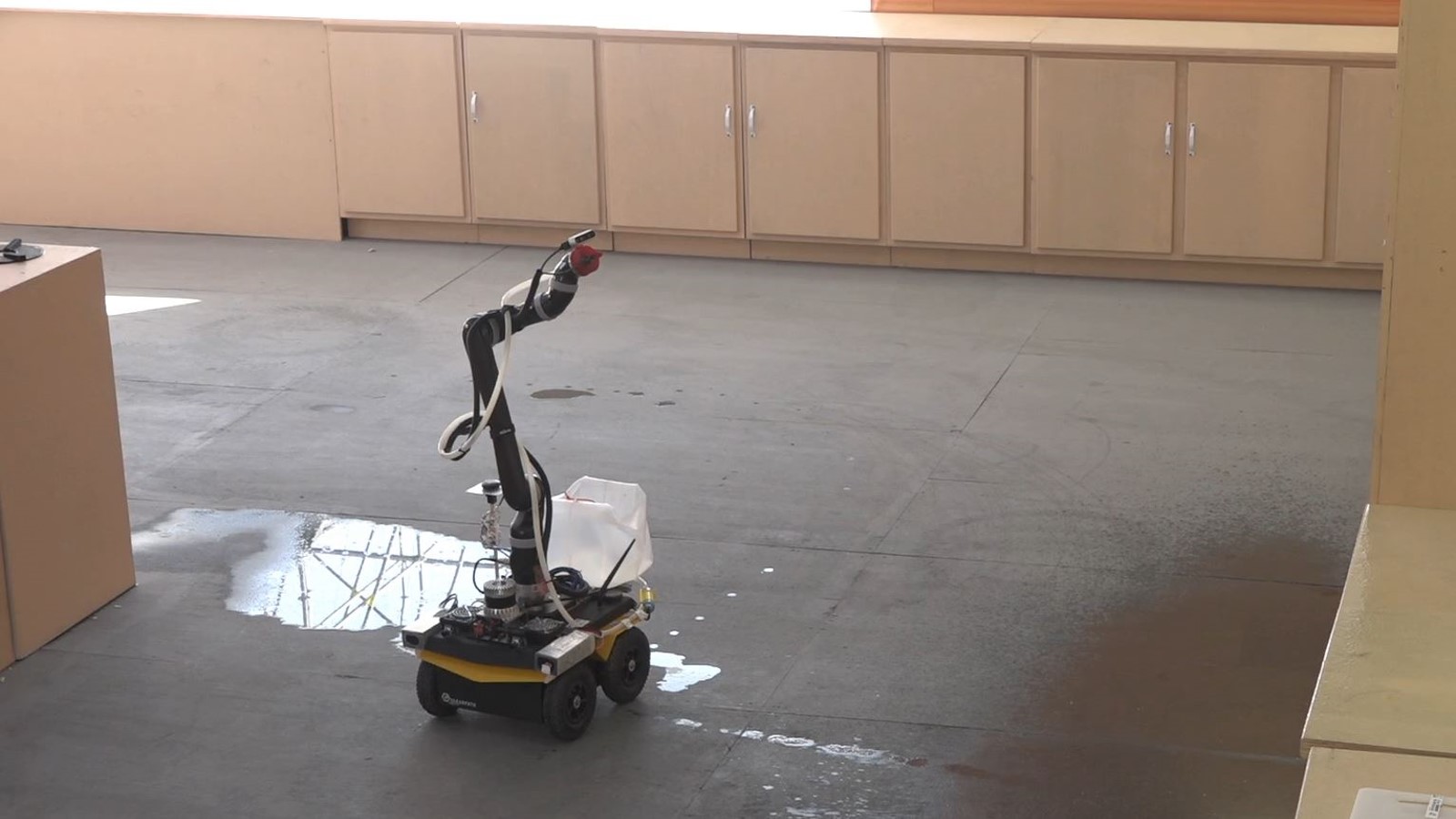}
    \includegraphics[width=0.32\textwidth]{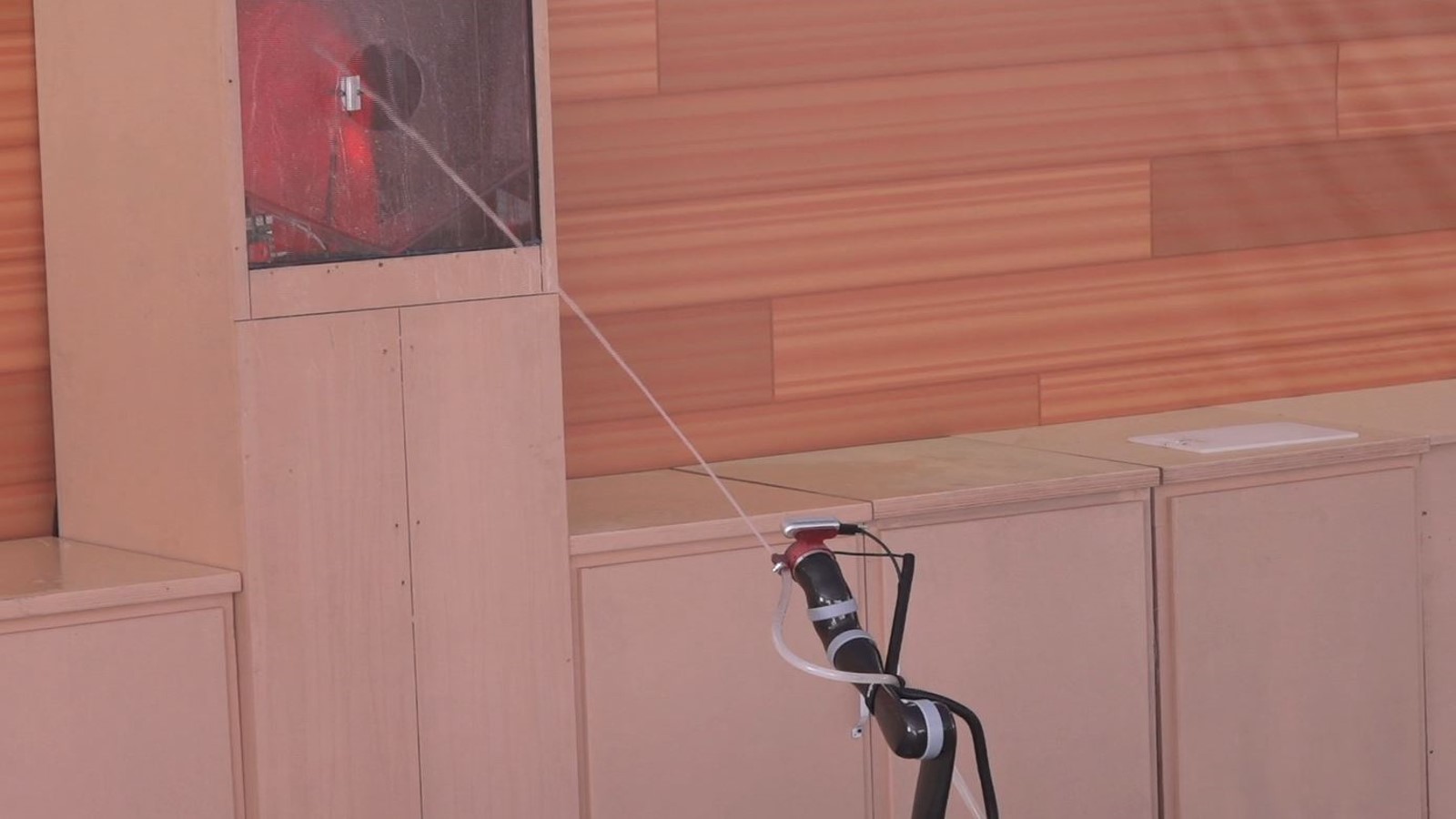}
    \caption{UGV navigating towards the target (left), performing heat source detection (center), and abating the detected heat (right).}
    \label{fig:moving}
\end{figure*} 
\begin{figure*}[!b]
    \centering
    \includegraphics[width=0.99\textwidth]{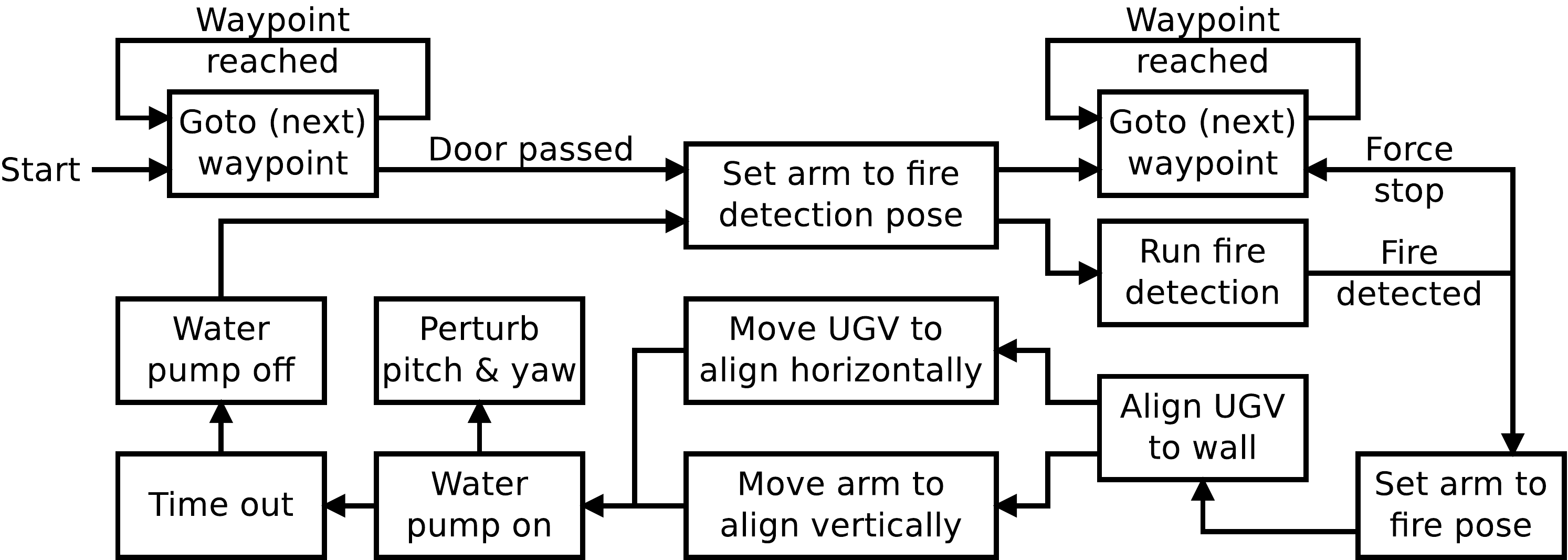}
	\caption{Overview of the state machine. The top part of the pipeline regards navigation and the bottom part corresponds to the fire extinguishing activity.}
    \label{fig:highLevelSM}
\end{figure*} 
In this section, we present results of the system based on the performance in the MBZIRC 2020 for Challenge No. 3 – Fire Fighting Inside a High-Rise Building shown in Fig.~\ref{fig:building_cad} (right). The key objective of this work is to navigate and localize the UGV, identify the heat source, and extinguish the heat source in the gound area of the building as shown in Fig.~\ref{fig:moving}. Since the test were performed on synthetic heat, by setting heat sources, the objective of this task was to shoot $1$ liter of water at the heat source target.
We developed a high-level execution policy to manage all the tasks described in the previous sections, using ROS. The execution policy, from here on referred as state machine, was developed to progress from one function to another based on the outcome of the current function. The state machine was developed using behavior engine FlexBE~\cite{2016:ICRA_Schillinger-etal}. It provides an extensive user interface to combine different states and execute them. The autonomy level for the transitions can be set by the user to control the system. Data can also pass from one state to the other. 

Each state in FlexBE corresponds to an action executed by the robotic system and based on the outcome of the action, it can then move to the appropriate state. It is a good platform for navigating a robot by providing waypoints, as well as for controlling robotic manipulator. To comply the feature with our system, we developed appropriate functions and algorithms using ROS for tasks such as activating and deactivating the water pump system, detecting the heat source using the thermal camera, and for aligning the robot to the target. For our system, all the transitions were fully autonomous. The transition of the system is based on the current state and on the outcome of the current state. Based on the outcome, we progress from one state to another. 
Figure~\ref{fig:highLevelSM} represents the high-level state machine on which the entire system runs.

\subsection{State Machine}~\label{sec:state_machine}
The state machine, shown in Fig.~\ref{fig:highLevelSM}, combines waypoint-based navigation with a fire detection and suppression procedure. 
There are two sets of waypoints: outdoor and indoor waypoints.
When navigating outdoors, the robot has its arm in a folded position, so it can accelerate and turn rapidly without the risk of slipping wheels or tripping over.
Indoor navigation is performed slowly with the arm in a raised position, so that the camera can inspect the wall surfaces completely.

During the competition the start location with respect to building was roughly known and the building was within scanning range of the LiDAR.
To avoid lengthy exploration of the entire arena, we exploited this a--priori knowledge and set the navigation waypoints manually.
In particular, we processed the 3D LiDAR scans and extracted the vertical planes of the building facade to obtain a 2D layout of the building.
This 2D layout was subsequently used to determine the positions of the waypoints the robot navigated through. The projected 2D layout also allows doors/gaps to be extracted easily.  

Inside the building, the vertical planes were assumed as the most appropriate structure to help guiding the robot.
In particular, the plane left on the robot was assumed and the waypoints were set so that the robot would move in clockwise fashion around the central piedestal of the building.
To ensure that the robot would inspect all of the wall surfaces, we extended the arm to position the end effector as high as possible. 

The distance from the walls was determined so that considering the FOV of the thermal camera, all of the inspected surfaces would be seen at least once.
Given the diagonal FOV of the thermal camera and distance from the wall we can assume that a fire will be easily visible if the camera height is kept fixed.
This simplifies the search problem as the arm can be held in a fixed position and the robot just navigates alongside the wall.
Hence we choose to set the waypoints to emulate a simple wall following strategy while searching the thermal source.
The two likely candidates for fire location were known a-priori: the outer walls of the room and a central pedestal.
This requires two possible loops inside the room, one with thermal camera pointing inwards while the other pointing thermal camera outwards.
For the first loop inside the room we chose to point the camera inwards and follow the walls.

While following the wall, we monitor the thermal camera output and if a temperature spot above certain threshold is reported, we stop the navigation. 
As the heat source is detected, we execute the fire extinguish steps, which are shown in the bottom part of \fig\ref{fig:highLevelSM}.
The arm is first set to fire abating position and the robot aligns itself to the wall.
The final aiming is then realised by moving the robot forward and backward and the nozzle up and down as described in Section~\ref{sec:aim}.
The pump is then set on and the nozzle yaw and pitch is perturbed in circular fashion to increase the water delivery area. 
After 20 seconds, the pump is switched off and we check if the fire is still present.
If the heat source is detected, we run the fire extinguish steps again.
If not, the robot proceeds to the next waypoint.

\begin{figure*}[!t]
    \centering
    \includegraphics[width=0.40\textwidth]{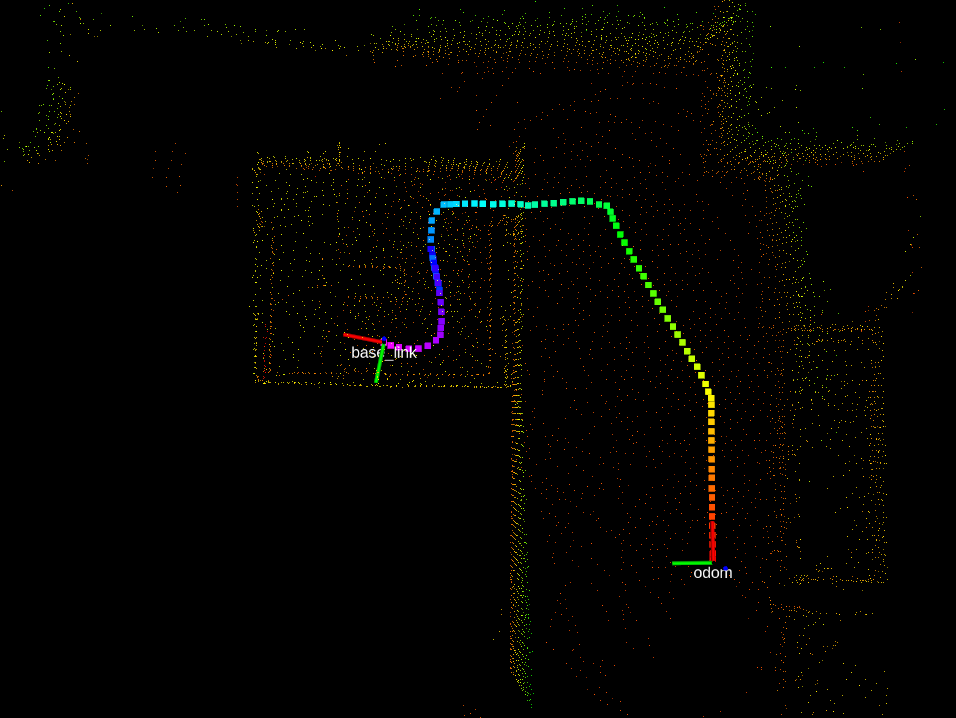}
    \includegraphics[width=0.40\textwidth]{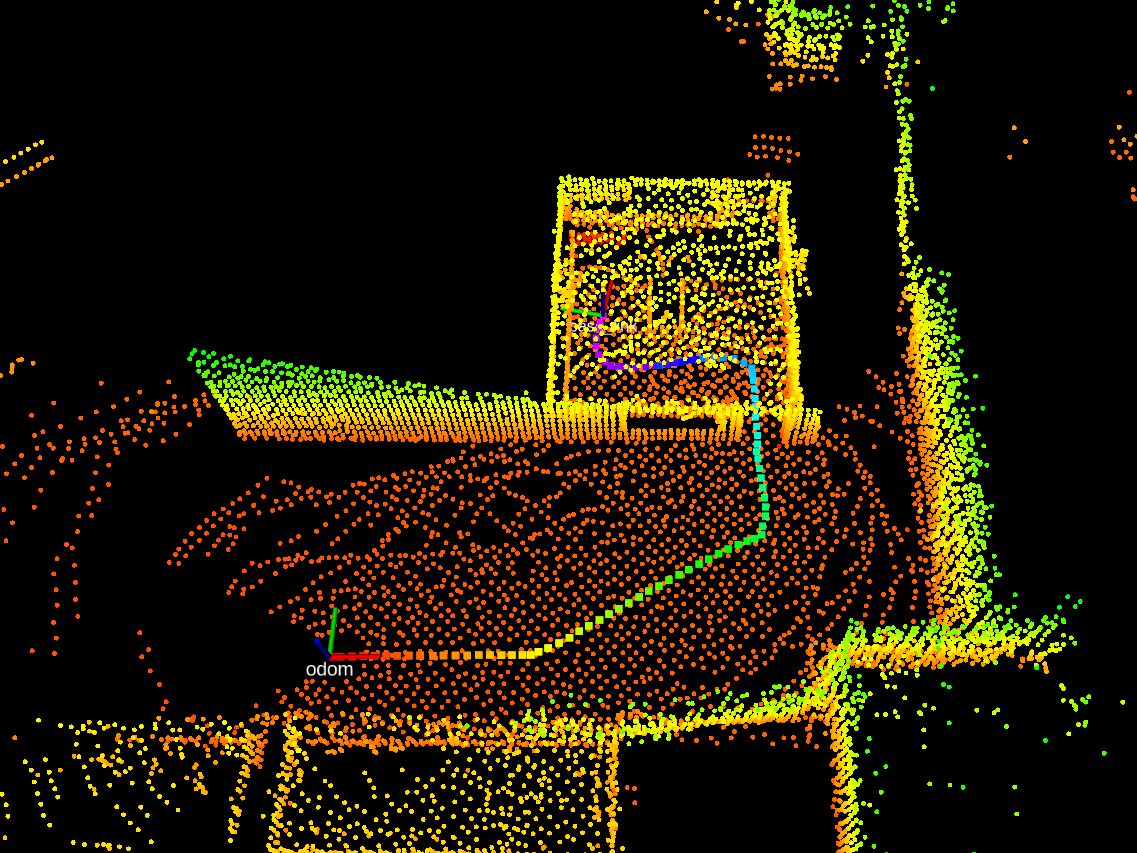}
    \caption{Trajectory of the UGV represented from two angles.}
    \label{fig:robTraj}
\end{figure*}
\subsection{Results}
Four experiments were performed as part of the MBZIRC 2020 Challenge No. 3 and the Grand Challenge.
While one can argue that the number of experiments is not sufficient to obtain statistically significant results, their aim is not to measure precision or accuracy, but to demonstrate reliability and robustness of the solution when deployed in a near-real scenario.
Moreover, the experiments are evaluated by an international jury, that does not allow to rule out cases, where the system failed for technical difficulties and of course numerous preliminary experiments of the system components as well as the entire system were done in the testing phase prior to the competition.
The evaluation of the experiments is based purely on the ability to deliver water through the narrow aperture of the fire targets.
The key questions of the experiments were: Is the proposed design made by off-the shelf components sufficient to accomplish the proposed task? More specifically, is the SLAM-based navigation and thermal source location pipeline precise enough to realize the fire abating by itself or do we need a--priori knowledge and specialized fire detection modules?

In all of the experiments, we let the robot reach the target locations with the arm extended in a way that in a case of precise localization, it would hit the simulated fire.
In the first run, we did not use an a--priori known map build by the robot 3D scanner and we simply set the waypoints using hand-measured positions.
In the subsequent runs, the waypoints were set using the map created by the 3D scanning system.
In the first and second experimental run, we did not use the final nozzle alignment specified in Section~\ref{sec:aim}.
In the third and fourth run, we used the fire detection modules to perform the final nozzle alignments. 
\begin{figure}[!t]
    \centering
    \includegraphics[width=0.7\linewidth]{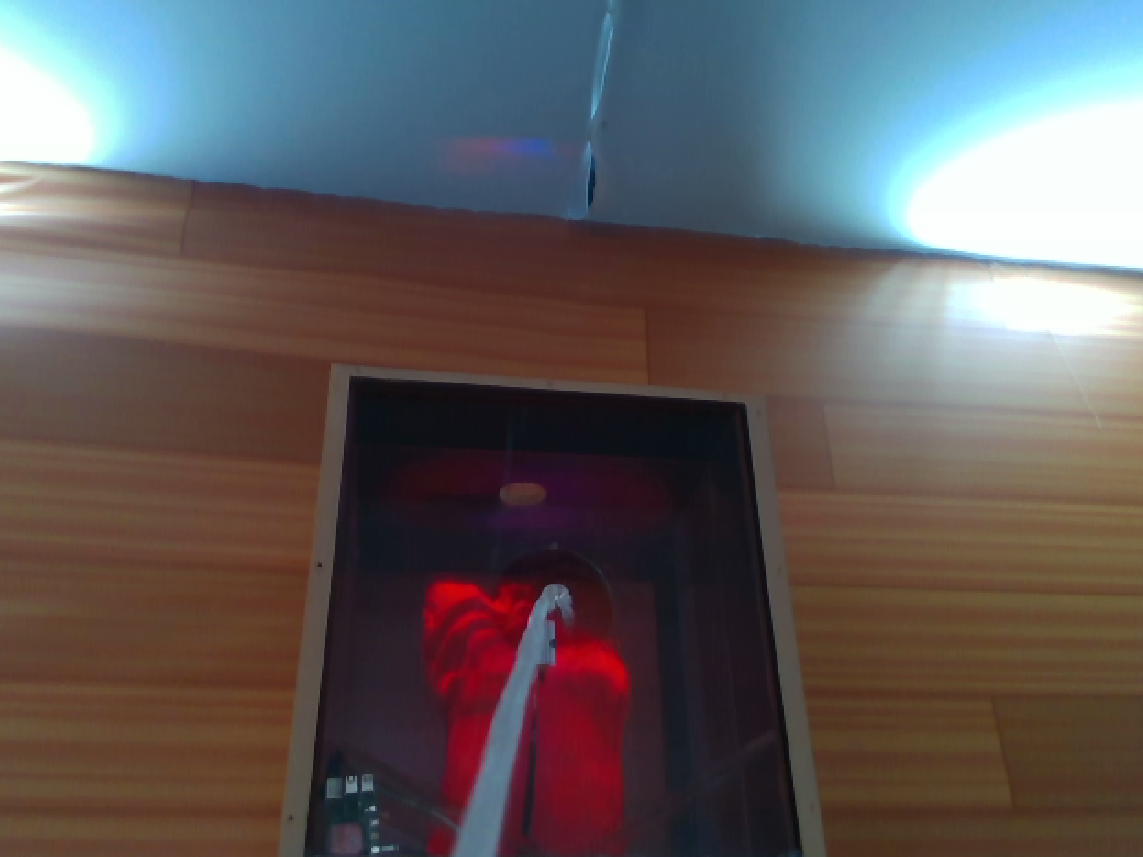}
    \caption{RGBD camera feed showing water being ejected to the heat source.}
    \label{fig:waterShot}
\end{figure} 
During first run, the hand-set waypoints were not precise enough to allow the path planning system to reliably pass through the door and the robot did not even enter the building. 
The second run used refined waypoints, and the robot passed through the door and navigated the building interiors without problems.
However, the localization drift prevented the robot to deliver water to the a--priori known artificial fire spots.
In the third and fourth runs, performed during the Grand Challenge round, the robot performed the final nozzle alignment using the methods specified in Section~\ref{sec:aim}.
In both of these runs, the artificial fire was detected correctly and water was delivered to it with ease. The average time to accomplish the overall mission was around 3 minutes. The attached multimedia material\footnote{\url{https://youtu.be/HCqQMEmkiN8}} further shows the successful performance of our system.

Figure~\ref{fig:robTraj} shows the trajectory of the UGV during the experiment.
The trajectory is shown in two different views, to visualize the arena in which the tests were performed.
The multicolored trail express the trajectory of the robot in the images.
To avoid tipping over backwards while carrying the heavy water payload the  UGV was operated at maximum velocity of $0.6$ m/s.
Loop-closure was not performed during the competition, but during manual tests the localization was sufficiently precise.
We empirically verified, due to the absence of ground truth, using the environment dimensions that similar localization errors are obtained compared to the one reported in original paper~\cite{8594299}.
The localization and mapping algorithm runs at $10$ Hz and consumes an average of $40\%$ of the overall CPU.

Figure~\ref{fig:thermal2} shows the feed from the thermal camera. The white blob in the image represents the heat source, which is successfully detected. The pixel location of the detected heat source in the thermal camera image is transformed to the RGB-D camera frame, using the approach explained in Section~\ref{sec:aim}. Once the transformation is performed, the heat source is localized, the UGV is aligned using the two step approach previously described, and the water is ejected on the heat source.
Figure~\ref{fig:waterShot} shows the image acquired from the RGB-D camera while the water is dispensed by the system to the detected heat source. Once the heat source is detected by thermal camera, and localized by the RGB-D camera, the water pump is turned on. The added cyclic motion of the arm increases the likelihood to shoot the target, while also covering greater area, which is beneficial in actual scenarios. The aim of the competition was to shoot $1$ liter of water towards the heat source. This goal was achieved successfully, but due to the size of the UGV, the water carried was not sufficient to extinguish the fire in one trip. The water container was refilled to extinguish the fire completely. Upon the refill, the robot performed all the tasks again successfully, with full autonomy.

\subsection{Lesson learned}
We discuss the research and technological challenges related to our approach and how the proposed solution can be transitioned from the MBZIRC test--bed to more complex real world urban firefighting scenarios.

First, the firefighting task definitely showed the importance to engage different researchers with cross-disciplinary competences in complementary research domains, such as localization, autonomous navigation, robotic arm control, fire detection, and hardware design to integrate the different robots parts and to obtain an end effector to house a thermal camera, stereo camera, and hose. The proposed solution can definitely inspire early roboticists to understand how the modules in the system are interconnected and work together to create an autonomous mobile robot solution for search and rescue scenarios.

Second, the system is not a monolithic solution to the firefighting problem.
Rather, it can be considered as a first step towards an efficient, rapid response system capable of early fire suppression.
The main problem of such highly task specialized systems -- the cost is actually mitigated by the versatile and modular design of our robot.
A mobile base with a manipulator could perform a variety of other tasks, such as delivery or facility inspection and the fire suppression could be achieved simply by exchanging the end effector~\cite{gripper}.
The presented system demonstrated that an off-the-shelf robotic platform, employed with customized state-of-the-art navigation, image processing and control methods can perform the fire suppression task reliably and efficiently.
Several customized parts needed were rapidly prototyped and 3D printed within few hours.
These aspects indicate that the field of mobile robotics is mature enough to deploy robots capable of autonomous, targeted fire suppression. 
Thus, mobile robots might soon compete with traditional sprinkler systems.

Finally, while the experiments indicate that the technology is ready to be deployed in buildings or small residential clusters, complex urban scenarios require more advanced, socially-aware navigation~\cite{social1,social2}, capable to deal with low visibility~\cite{fog}.
These capabilities have not been addressed in the proposed approach and will require further future investigation.
On the other hand, the proposed approach for robot localization provides direct feedback control to the robot event transitioning from outdoor to indoor settings.
The water tank and automated water pump designs for fire abatement are simple, inexpensive, and effective.
Similar mechanical and circuitry designs and concepts can be used by private or government organization as firefighting solution.
To demonstrate the versatility of the approach, we open source our mechanical and electrical designs and navigation solution~\cite{github}.

%% file: sections/conclusion.tex
\section{Conclusion}\label{sec:conclusion}

In this work, we developed a mobile manipulator equipped with a specialized end--effector to detect heat sources in indoor environments and abate them.
The challenges included autonomous navigation of the mobile platform, design heat source identification, and control of a water pump system such to obtain fire abatement.
\begin{figure}[!b]
    \centering
    \includegraphics[width=0.8\linewidth]{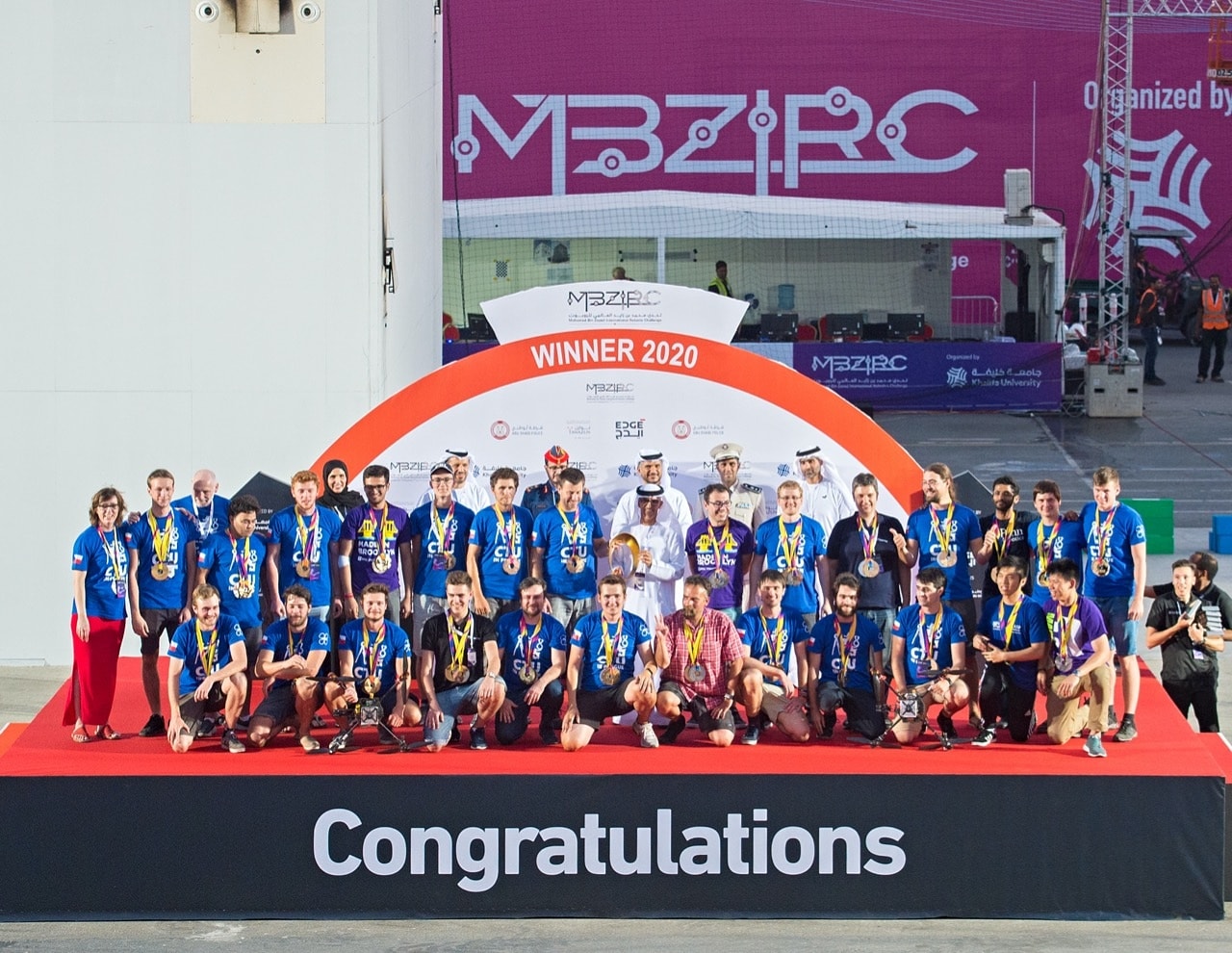}
    \caption{Our team during the ceremony celebrating the first place in the Grand Challenge.}
    \label{fig:mbzirc_competition}
\end{figure}
The presented results show the effectiveness of our approach and its applicability in real firefighting scenarios.
We believe that the proposed solution can be considered a first step toward the transitioning ground robotic solutions to realistic rapid fire suppression systems and search and rescue scenarios. The system has been designed as part of a solution in the MBZIRC 2020 for the  Challenge No. 3 – Fire Fighting Inside a High-Rise Building and for the Grand Challenge. The proposed solution was instrumental and one of the key elements to win the Grand Challenge in the MBZIRC competition as captured in Fig.~\ref{fig:mbzirc_competition}. It was the smallest UGV solution proposed among the competing teams and scored during the Grand Challenge the highest number of points among all UGV firefighting solutions presented by the 17 teams admitted to the final competition.

Future works will consist the ability to increase the autonomy and localization by detecting semantic cues in the environments, tracking heat sources concurrently while moving the UGV and the arm, and incorporation of exchangeable end effectors with different properties. Although developed, for the specific needs of the competition, the system has shown to be a promising solution to be employed in the future in real and complex urban firefighting scenarios and has the potential to be commercialized as well by our team members.

\section*{Acknowledgements}
The outstanding results of the MBZIRC project would not have been achieved without the help and collaboration of each member of our team, comprising people from the Czech Technical University in Prague, the University of Pennsylvania, and New York University (see Fig.~\ref{fig:mbzirc_competition}).
This work was supported by the Mohamed Bin Zayed International Robotics Challenge through Khalifa University that supported 17 selected teams for the MBZIRC competition.

%% file: main.bbl
\begin{thebibliography}{}

\bibitem[Ahrens, 2017]{sprinklers}
Ahrens, M. (2017).
\newblock {US} experience with sprinklers.
\newblock Technical report.

\bibitem[{Altaf} et~al., 2007]{4381341}
{Altaf}, K., {Akbar}, A., and {Ijaz}, B. (2007).
\newblock Design and construction of an autonomous fire fighting robot.
\newblock In {\em 2007 International Conference on Information and Emerging
  Technologies}, pages 1--5.

\bibitem[{Anantha Raj} and {Srivani}, 2018]{Raj}
{Anantha Raj}, P. and {Srivani}, M. (2018).
\newblock Internet of robotic things based autonomous fire fighting mobile
  robot.
\newblock In {\em 2018 IEEE International Conference on Computational
  Intelligence and Computing Research (ICCIC)}, pages 1--4.

\bibitem[{Banerjee} et~al., 2018]{8383972}
{Banerjee}, D., {Yu}, K., and {Aggarwal}, G. (2018).
\newblock Image rectification software test automation using a robotic arm.
\newblock {\em IEEE Access}, 6:34075--34085.

\bibitem[Broughton et~al., 2020]{fog}
Broughton, G., Majer, F., et~al. (2020).
\newblock Learning to see through the haze: Multi-sensor learning-fusion system
  for vulnerable traffic participant detection in fog.
\newblock {\em Robotics and Autonomous Systems}.
\newblock to appear.

\bibitem[{Chen} et~al., 2018]{8686979}
{Chen}, X., {Zhao}, B., and {Gao}, X. (2018).
\newblock Noninvasive brain-computer interface based high-level control of a
  robotic arm for pick and place tasks.
\newblock In {\em 2018 14th International Conference on Natural Computation,
  Fuzzy Systems and Knowledge Discovery (ICNC-FSKD)}, pages 1193--1197.

\bibitem[{Chitta} et~al., 2012]{6174325}
{Chitta}, S., {Sucan}, I., and {Cousins}, S. (2012).
\newblock Moveit! [ros topics].
\newblock {\em IEEE Robotics \& Automation Magazine}, 19(1):18--19.
\newblock \url{https://moveit.ros.org/}.

\bibitem[{Deng} et~al., 2017]{mobile_MoveIt}
{Deng}, H., {Xiong}, J., and {Xia}, Z. (2017).
\newblock Mobile manipulation task simulation using ros with moveit.
\newblock In {\em 2017 IEEE International Conference on Real-time Computing and
  Robotics (RCAR)}, pages 612--616.

\bibitem[Ebadi et~al., 2020]{LAMP}
Ebadi, K., Chang, Y., Palieri, M., Stephens, A., Hatteland, A., Heiden, E.,
  Thakur, A., Morrell, B., Wood, S., Carlone, L., et~al. (2020).
\newblock Lamp: Large-scale autonomous mapping and positioning for exploration
  of perceptually-degraded subterranean environments.
\newblock In {\em IEEE International Conference on Robotics and Automation
  (ICRA)}.

\bibitem[FEMA, 2020a]{residentialBuldingStats}
FEMA (2020a).
\newblock Fire estimate summary.
\newblock
  \url{https://www.usfa.fema.gov/downloads/pdf/statistics/res_bldg_fire_estimates.pdf}.

\bibitem[FEMA, 2020b]{nonResidentialBuldingStats}
FEMA (2020b).
\newblock Fire estimate summary.
\newblock
  \url{https://www.usfa.fema.gov/downloads/pdf/statistics/nonres_bldg_fire_estimates.pdf}.

\bibitem[{Hassanein} et~al., 2015]{7251507}
{Hassanein}, A., {Elhawary}, M., {Jaber}, N., and {El-Abd}, M. (2015).
\newblock An autonomous firefighting robot.
\newblock In {\em 2015 International Conference on Advanced Robotics (ICAR)},
  pages 530--535.

\bibitem[Khoon et~al., 2012]{KHOON20121145}
Khoon, T.~N., Sebastian, P., and Saman, A. B.~S. (2012).
\newblock Autonomous fire fighting mobile platform.
\newblock {\em Procedia Engineering}, 41:1145 -- 1153.
\newblock International Symposium on Robotics and Intelligent Sensors 2012
  (IRIS 2012).

\bibitem[{Kim} et~al., 2009]{5353970}
{Kim}, Y., {Kim}, Y., {Lee}, S., {Kang}, J., and {An}, J. (2009).
\newblock Portable fire evacuation guide robot system.
\newblock In {\em 2009 IEEE/RSJ International Conference on Intelligent Robots
  and Systems}, pages 2789--2794.

\bibitem[Kong et~al., 2004]{gripper}
Kong, F.-k., Diao, Y.-f., and Yang, E.-x. (2004).
\newblock Research on automatic robot manipulator exchange.
\newblock {\em Mechanical Engineer}, 5.

\bibitem[Kucner et~al., 2017]{social1}
Kucner, T.~P., Magnusson, M., Schaffernicht, E., Bennetts, V.~H., and
  Lilienthal, A.~J. (2017).
\newblock Enabling flow awareness for mobile robots in partially observable
  environments.
\newblock {\em IEEE Robotics and Automation Letters}, 2(2):1093--1100.

\bibitem[{Kuffner} and {LaValle}, 2000]{844730}
{Kuffner}, J.~J. and {LaValle}, S.~M. (2000).
\newblock Rrt-connect: An efficient approach to single-query path planning.
\newblock In {\em IEEE International Conference on Robotics and Automation},
  pages 995--1001.

\bibitem[{Lee} et~al., 2018]{8384639}
{Lee}, J., {Li}, W., {Shen}, J., and {Chuang}, C. (2018).
\newblock Multi-robotic arms automated production line.
\newblock In {\em 2018 4th International Conference on Control, Automation and
  Robotics (ICCAR)}, pages 26--30.

\bibitem[{Maddukuri} et~al., 2016]{7977097}
{Maddukuri}, S. V. P.~K., {Renduchintala}, U.~K., {Visvakumar}, A., {Pang}, C.,
  and {Mittapally}, S.~K. (2016).
\newblock A low cost sensor based autonomous and semi-autonomous fire-fighting
  squad robot.
\newblock In {\em 2016 Sixth International Symposium on Embedded Computing and
  System Design (ISED)}, pages 279--283.

\bibitem[Minkina and Dudzik, 2009]{minkina2009infrared}
Minkina, W. and Dudzik, S. (2009).
\newblock {\em Infrared thermography: errors and uncertainties}.
\newblock John Wiley \& Sons.

\bibitem[Quigley et~al., 2009]{ros}
Quigley, M., Conley, K., Gerkey, B.~P., Faust, J., Foote, T., Leibs, J.,
  Wheeler, R., and Ng, A.~Y. (2009).
\newblock Ros: an open-source robot operating system.
\newblock In {\em ICRA Workshop on Open Source Software}.

\bibitem[{Rahman} et~al., 2019]{9087500}
{Rahman}, R., {Rahman}, M.~S., and {Bhuiyan}, J.~R. (2019).
\newblock Joystick controlled industrial robotic system with robotic arm.
\newblock In {\em 2019 IEEE International Conference on Robotics, Automation,
  Artificial-intelligence and Internet-of-Things (RAAICON)}, pages 31--34.

\bibitem[{Saraee} et~al., 2017]{8007498}
{Saraee}, E., {Joshi}, A., and {Betke}, M. (2017).
\newblock A therapeutic robotic system for the upper body based on the proficio
  robotic arm.
\newblock In {\em 2017 International Conference on Virtual Rehabilitation
  (ICVR)}, pages 1--2.

\bibitem[Schillinger et~al., 2016]{2016:ICRA_Schillinger-etal}
Schillinger, P., Kohlbrecher, S., and von Stryk, O. (2016).
\newblock Human-robot collaborative high-level control with application to
  rescue robotics.
\newblock In {\em Proc. IEEE Int. Conf. on Robotics and Automation (ICRA)},
  pages 2796--2802.

\bibitem[{Shan} and {Englot}, 2018]{8594299}
{Shan}, T. and {Englot}, B. (2018).
\newblock Lego-loam: Lightweight and ground-optimized lidar odometry and
  mapping on variable terrain.
\newblock In {\em 2018 IEEE/RSJ International Conference on Intelligent Robots
  and Systems (IROS)}, pages 4758--4765.

\bibitem[{Tamilselvi} et~al., 2018]{8399114}
{Tamilselvi}, R., {Merline}, A., {Beham}, M.~P., {Anand}, R.~V., {Karthik},
  M.~S., and {Uthayakumar}, R.~H. (2018).
\newblock Emg activated robotic arm for amputees.
\newblock In {\em 2018 2nd International Conference on Inventive Systems and
  Control (ICISC)}, pages 456--461.

\bibitem[Tardioli et~al., 2019]{doi:10.1002/rob.21871}
Tardioli, D., Riazuelo, L., Sicignano, D., Rizzo, C., Lera, F., Villarroel,
  J.~L., and Montano, L. (2019).
\newblock Ground robotics in tunnels: Keys and lessons learned after 10 years
  of research and experiments.
\newblock {\em Journal of Field Robotics}, 36(6):1074--1101.

\bibitem[Thakur et~al., 2020]{github}
Thakur, D., Loianno, G., {\v S}tibinger, P., and Krajn{\'i}k, T. (2020).
\newblock Source codes: Mbzirc firefighting challenge upenn-ctu-nyu solution.

\bibitem[{Thrun} et~al., 2003]{1242260}
{Thrun}, S., {Hahnel}, D., {Ferguson}, D., {Montemerlo}, M., {Triebel}, R.,
  {Burgard}, W., {Baker}, C., {Omohundro}, Z., {Thayer}, S., and {Whittaker},
  W. (2003).
\newblock A system for volumetric robotic mapping of abandoned mines.
\newblock In {\em 2003 IEEE International Conference on Robotics and Automation
  (Cat. No.03CH37422)}, volume~3, pages 4270--4275 vol.3.

\bibitem[{Uaday} et~al., 2019]{Zaman}
{Uaday}, M.~A., {Nazmul Islam Shuzan}, M., {Shanewaze}, S., {Rakib}, R.~I., and
  {Zaman}, H.~U. (2019).
\newblock The design of a novel multi-purpose fire fighting robot with video
  streaming capability.
\newblock In {\em 2019 IEEE 5th International Conference for Convergence in
  Technology (I2CT)}, pages 1--5.

\bibitem[Vintr~et al., 2020]{social2}
Vintr~et al., T.~. (2020).
\newblock Natural criteria for comparison of pedestrian flow forecasting
  models.
\newblock In {\em IEEE/RSJ International Conference on Intelligent Robots and
  Systems (IROS)}. IEEE.
\newblock to appear.

\bibitem[{Weng} and {Chen}, 2017]{8274801}
{Weng}, C. and {Chen}, I. (2017).
\newblock The task-level evaluation model for a flexible assembly task with an
  industrial dual-arm robot.
\newblock In {\em 2017 IEEE International Conference on Cybernetics and
  Intelligent Systems (CIS) and IEEE Conference on Robotics, Automation and
  Mechatronics (RAM)}, pages 356--360.

\bibitem[{Ye} et~al., 2019]{8996761}
{Ye}, Z., {Su}, F., {Zhang}, Q., and {Wan}, L. (2019).
\newblock Intelligent fire-fighting robot based on stm32.
\newblock In {\em 2019 Chinese Automation Congress (CAC)}, pages 3369--3373.

\bibitem[Zhang and Singh, 2014]{Zhang-2014-7903}
Zhang, J. and Singh, S. (2014).
\newblock Loam: Lidar odometry and mapping in real-time.
\newblock In {\em Proceedings of Robotics: Science and Systems Conference}.

\end{thebibliography}
